\documentclass[conference,a4paper]{IEEEtran}
\IEEEoverridecommandlockouts

\usepackage[hidelinks]{hyperref}
\usepackage[cmex10]{amsmath}
\usepackage{amssymb,amsfonts}
\usepackage{dblfloatfix}

\usepackage{soul}
\usepackage{color}

\usepackage[ruled,vlined]{algorithm2e}
\usepackage{graphicx}

\usepackage{booktabs}
\usepackage{siunitx}
\usepackage[numbers,compress]{natbib}
\usepackage{texnames}
\usepackage{bm,bbm}
\usepackage{orcidlink}
\usepackage{amsmath}
\usepackage{array}
\usepackage{multirow}
\usepackage{colortbl}
\usepackage{booktabs}
\usepackage{xcolor}

\usepackage{tabularray}

\usepackage{microtype}

\usepackage{bbold}

\begin{document}
\title{Detection of Bark Beetle Attacks using Hyperspectral PRISMA Data and Few-Shot Learning
\thanks{This activity has been supported by Italian Space Agency under the call PRISMA SCIENZA ASI DC-UOT-2019-061, financing agreement ASI N. 2022-1-U.0 “AFORISMA: Machine learning for forest cover analysis with hyperspectral data from the PRISMA mission to support the National Forest Inventory”.}
}

\author{    \IEEEauthorblockN{Mattia Ferrari}
	\IEEEauthorblockA{\textit{University of Trento}\\ 38122 Trento, Italy\\ mattia.ferrari-2@unitn.it}
	\and
        \IEEEauthorblockN{Giancarlo Papitto}
	\IEEEauthorblockA{\textit{Arma dei Carabinieri}\\ 00187 Roma, Italy\\ giancarlo.papitto@carabinieri.it}
	\and
        \IEEEauthorblockN{Giorgio Deligios}
	\IEEEauthorblockA{\textit{Arma dei Carabinieri}\\ 00187 Roma, Italy\\ giorgio.deligios@carabinieri.it}
	\and
	\IEEEauthorblockN{Lorenzo Bruzzone}
	\IEEEauthorblockA{\textit{University of Trento}\\ 38122 Trento, Italy\\ lorenzo.bruzzone@unitn.it}
}

\maketitle
\begin{abstract}
Bark beetle infestations represent a serious challenge for maintaining the health of coniferous forests. This paper proposes a few-shot learning approach leveraging contrastive learning to detect bark beetle infestations using satellite PRISMA hyperspectral data. The methodology is based on a contrastive learning framework to pre-train a one-dimensional CNN encoder, enabling the extraction of robust feature representations from hyperspectral data. These extracted features are subsequently utilized as input to support vector regression estimators, one for each class, trained on few labeled samples to estimate the proportions of healthy, attacked by bark beetle, and dead trees for each pixel. Experiments on the area of study in the Dolomites show  that our method outperforms the use of original PRISMA spectral bands and of Sentinel-2 data. The results indicate that PRISMA hyperspectral data combined with few-shot learning offers significant advantages for forest health monitoring.
\end{abstract}

\begin{IEEEkeywords} 
Deep learning, hyperspectral, bark beetle, PRISMA, remote sensing.
\end{IEEEkeywords}

\section{Introduction}

Bark beetle infestations pose a significant threat to coniferous health, where they contribute to extensive tree mortality and ecosystem degradation \cite{hlasny2021bark}. The ecological and environmental impacts of bark beetle outbreaks emphasize the critical need for precise detection and monitoring methods to guide effective forest management and mitigation strategies \cite{edburg2012cascading}.

In recent years, multispectral data acquired by spaceborne sensors, combined with machine learning (ML) techniques, have become essential tools for the detection of bark beetle infestations. Despite challenges related to the spatial resolution that does not allow to work at the individual tree level, they facilitate the analysis of extensive and often inaccessible areas \cite{marvasti2023early} \cite{kautz2024early}.
For example, Sentinel-2 satellite imagery, with its 13 spectral bands, has demonstrated effectiveness in detecting bark beetle infestations by enabling the identification of vegetation stress mainly through strategies that careful select and combine spectral bands. However, the relatively broad spectral bandwidth of Sentinel-2 limits its capacity to capture subtle variations in vegetation health, which may reduce the sensitivity required for detection.

In contrast, hyperspectral remote sensing, which captures an almost continuous spectrum across hundreds of narrow bands, provides superior spectral resolution, enabling the detection of finer variations in vegetation health. This increased spectral detail facilitates the identification of subtle stress indicators that are often undetectable with multispectral sensors, thereby enhancing the sensitivity of early detection. Although hyperspectral data has been used for bark beetle detection primarily from UAVs and aerial platforms, recent advancements in satellite-based hyperspectral sensors, such as PRISMA \cite{guarini2017overview}, offer significant potential for improving forest health monitoring. PRISMA acquires hyperspectral data across 234 bands with a spatial resolution of 30 meters, thus representing a notable potential advancement in the capability to monitor forest ecosystems at larger scales.

In the literature, traditional ML approaches, such as random forests and support vector machines, are frequently utilized for detecting bark beetle infestations, often in conjunction with feature engineering techniques. A common approach involves the use of vegetation indices, which are mathematical combinations of spectral bands designed to highlight specific aspects of vegetation health. These indices can be either selected in their entirety or through an automatic selection of the most relevant ones \cite{dalponte2022mapping}. Additionally, these indices are often combined with raw spectral bands \cite{marvasti2023early}.
In recent years, deep learning (DL) models have gained significant success in various classification and regression tasks and have been effectively applied to bark beetle detection, primarily using multispectral and hyperspectral UAV data and Convolutional Neural Networks (CNNs), yielding promising results \cite{turkulainen2023comparison}. However, there has been limited exploration of using DL methodologies with hyperspectral satellite data for bark beetle detection.

A key challenge in applying ML to remote sensing data is acquiring a sufficient number of reliable ground truth samples, as the effectiveness of models is significantly compromised when trained on a limited amount of reference data. This limitation is particularly pronounced when attempting to exploit the full potential of hyperspectral images, due to the high dimensionality of the data, which further increases the need for a large number of labeled samples. Collecting these samples typically involves extensive and time-intensive fieldwork, requiring data that are both accurate and representative. This is especially challenging when studying complex and dynamic phenomena like bark beetle infestations, which vary significantly across space and time. Additionally, manually identifying and annotating individual trees in dense forests is prone to errors and inconsistencies, which complicates the development of reliable training datasets. Consequently, ML models designed for detecting bark beetle infestations are often trained on a relatively small number of ground truth samples, underscoring the need for methodologies capable of delivering accurate results with minimal labeled data.

To address the challenge of limited labeled datasets, few-shot learning (FSL) \cite{sun2021research} techniques have gained significant attention. FSL methods are designed to generalize from few labeled samples, making them particularly effective in scenarios with scarce ground truth data. Among the various FSL frameworks, contrastive learning \cite{lim2023scl} has proven to be one of the most effective. This approach learns discriminative feature representations by contrasting similar and dissimilar sample pairs, facilitating the model ability to capture underlying data relationships. Contrastive learning can be adapted for few-shot scenarios by using the limited labeled data to form meaningful positive and negative pairs. By emphasizing the relationships within these few samples, the model can be fine-tuned to generalize effectively to new unseen data, achieving robust performance even with minimal supervision.

In this paper, we propose a few-shot learning FSL method leveraging contrastive learning to detect bark beetle infestations using PRISMA hyperspectral data at the sub-pixel level. The method utilizes a contrastively trained model, fine-tuned with a limited number of labeled samples, to extract relevant features. These features are then used as input to support vector regression (SVR) estimators to quantify the proportions of healthy, affected, and dead trees relative to the total number of trees within each image pixel. By leveraging the rich spectral information provided by hyperspectral data, our approach effectively addresses the challenges posed by the availability of a small number of ground truth samples, while partially mitigating the limitations associated with the relatively low spatial resolution of PRISMA. To assess the performance of our method, we compare it with the use of only the spectral bands from both PRISMA and Sentinel-2 data maintaining the same classifier. This comparison highlights the effectiveness of the proposed method.
\section{Methodology}

\begin{figure}
  \includegraphics[width=\columnwidth]{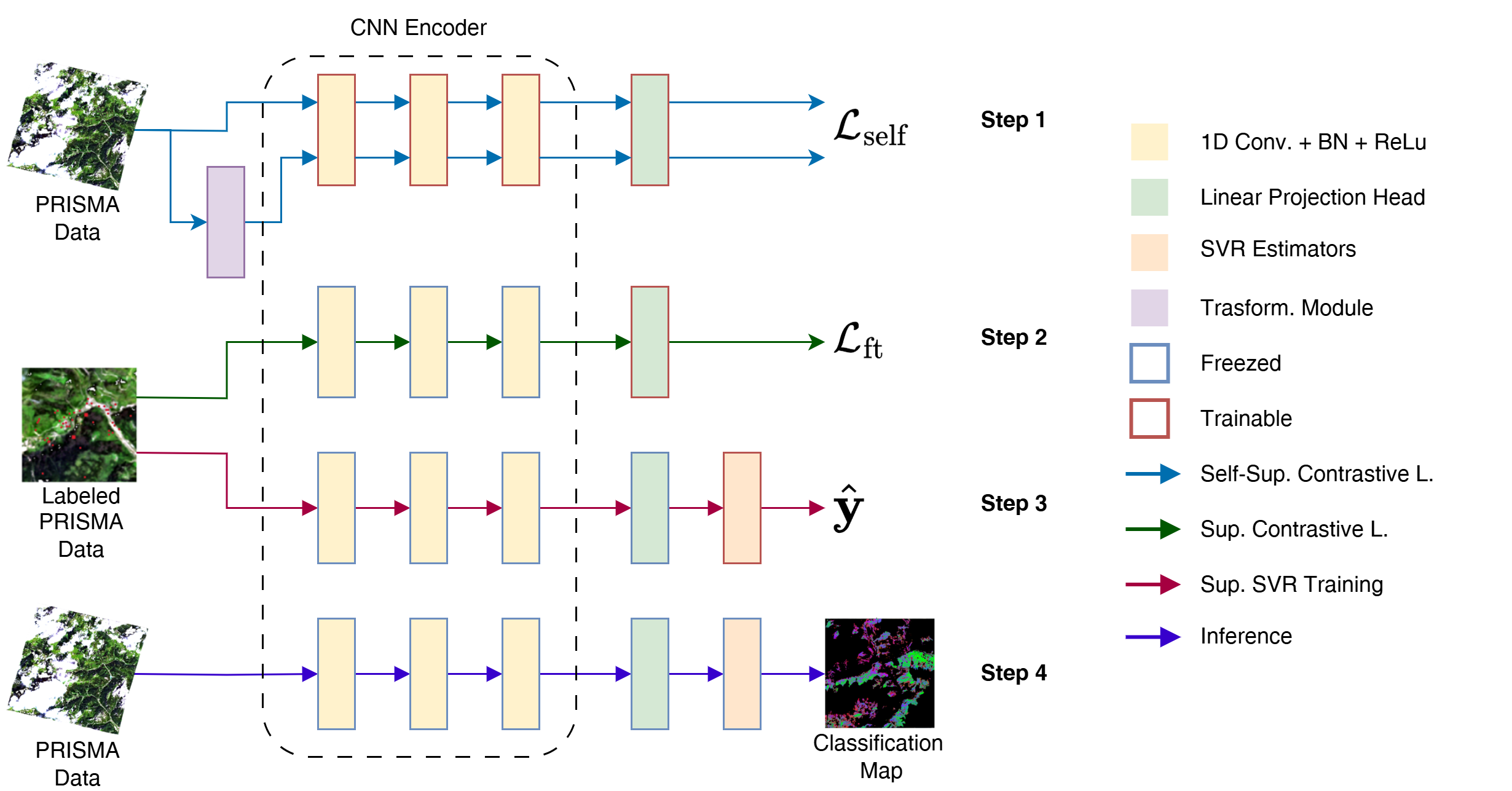}
  \caption{Scheme of the proposed method.}
  \label{fig:scheme}
\end{figure}

The proposed method, illustrated in Figure \ref{fig:scheme}, incorporates a contrastive learning framework to pre-train the model on unlabeled hyperspectral data, allowing the learning of robust feature representations. In the subsequent fine-tuning phase, the final layer of the model is adapted using the limited labeled samples to optimize performance for the specific task. Finally,  support vector regressor (SVR)  estimators, one for each class, are trained using the features extracted from the fine-tuned contrastive learned model to predict the infestation severity, expressed as the abundance of healthy, affected and dead trees within each pixel. The primary advantage of this approach is its ability to compress the spectral signature information using minimal labeled data, specifically for the sub-pixel detection of bark beetle infestations. In this context, the use of SVR estimators is motivated by their superior ability to provide accurate estimations with limited data, as well as their general tendency to mitigate overfitting compared to fully connected layers.

The contrastive learned encoder $e(\cdot)$ consists of a one-dimensional CNN that takes the spectral signature of a hyperspectral pixel, $\mathbf{x}$, as input. The model is composed of three convolutional layers with decreasing kernel sizes (7, 5, 3) and increasing numbers of channels (32, 64, 128), each followed by batch normalization layers and ReLU activation functions. The output $e(\mathbf{x})$ is the latent features representation of $\mathbf{x}$ with a dimensionality of 128. This latent representation is then passed through a linear projection head, $h(\cdot)$, which maps it to a 16-dimensional embedding space, yielding a compressed representation of the input spectral signature: $\mathbf{z} = h(e(\mathbf{x}))$. These 16 features are subsequently fed into three independent SVRs, each responsible for predicting one of the target labels. Finally, the predicted values are normalized to ensure compliance with the non-negativity and sum-to-one constraints.

To train $e$, we take as input a batch of $B$ spectral signature of samples, and we forward it through the network, obtaining its embedding representation. Then we transform each sample in the batch $\mathbf{x}_{1},\;...\;,\mathbf{x}_{b},\;...\;,\mathbf{x}_{B}$ scaling and noising it and then applying a magnitude warping \cite{um2017data}. This process is expressed in the following equation:
\begin{equation} \label{eq:warping}
\mathbf{x}^{'} = \boldsymbol{\beta} (\alpha \mathbf{x} + \boldsymbol{\mathcal{N}})
\end{equation}
where $\boldsymbol{\mathcal{N}}$ is a zero-mean random Gaussian noise with standard deviation $\sigma_1$, $\alpha$ a scaling factor and the $\boldsymbol{\beta}$ parameter sequence is created by interpolating a cubic spline $S(\mathbf{u})$ with $N$ knots. Each knot $u_n$ in $\mathbf{u}$ is taken from the Gaussian distribution $\mathcal{N}(1,{\sigma_2}^2)$.

The batch of samples transformed $\mathbf{x}'_{1},\;...\;,\mathbf{x}'_{b},\;...\;,\mathbf{x}'_{B}$ is then forwarded to the network, obtaining a batch of $\mathbf{z}_b^{'}$. 
Then, the two embedded batches are concatenated and the loss of this process $\mathcal{L}_{\text{self}}$ is computed using the SimCLR contrastive loss \cite{chen2020simple}, expressed as follows:

\begin{equation}
\mathcal{L}_{\text{self}} = - \frac{1}{2B} \sum_{i=1}^{2B} \log \frac{\exp\left(\frac{\text{sim}(\mathbf{z}_i, \mathbf{z}_i^+)}{\tau}\right)}{\sum_{j=1}^{2B} \mathbb{1}_{[j\neq i]} \exp\left(\frac{\text{sim}(\mathbf{z}_i, \mathbf{z}_j)}{\tau}\right)}
\end{equation}
where $\mathbf{z}_i$ is the representation of the anchor sample, $\mathbf{z}_i^+$ the positive sample associated to $\mathbf{z}_i$, that is equal to ${\mathbf{z}_{i+B}}$ if $i < B$ or ${\mathbf{z}_{i-B}}$ if, $i \ge B$, $\tau$ is a temperature parameter that controls the concentration of the distribution and the $\text{sim}$ function is defined as cosine similarity formulated as follows:
\begin{equation}
\text{sim}(\mathbf{u}, \mathbf{v}) = \frac{\mathbf{u}^{\top} \cdot \mathbf{v}}{\|\mathbf{u}\| \|\mathbf{v}\|}
\end{equation}

After the self-supervised training of $h$ and $e$, we freeze the contrastive learned encoder $e$, and apply a supervised fine-tuning to the trained model $h(e(\cdot))$ on all the $N$ labeled samples $\mathbf{x}_i^l$ with label vector $\mathbf{y}_i$. The fine-tuning loss $\mathcal{L}_{\text{ft}}$ is computed using the following equation:

\begin{equation}
\mathcal{L}_{\text{ft}}=-\frac{1}{N}\sum_{i=1}^{N}\text{log}\frac{\sum_{k=1}^{K}\text{exp} (\frac{\text{sim}(\textbf{z}_i,\textbf{z}_k^*)}{\tau})}{\sum_{j=1}^N \exp\left(\frac{\text{sim}(\mathbf{z}_i, \mathbf{z}_j)}{\tau}\right)}
\end{equation}
where $\textbf{z}_k^*$ are all the $K$ embeddings for which the following condition is respected: $||\textbf{y}_i-\textbf{y}_k||<\lambda$ with $\lambda$ empirically set equal to 0.6.

This process allows fine-tuning the latent of the extracted features to optimize their alignment with the bark beetle detection task, while mitigating the reliance on a limited set of labeled samples. The final step of the training procedure involves training the independent SVRs using the embeddings  $\textbf{z}_i^l$ of the labeled samples $\mathbf{x}_i^l$. Once the SVRs are trained, they enable the estimation of abundances for new, unseen data. Specifically, embeddings are extracted from the new data via the contrastive encoder and the projection layer, and their corresponding abundances are then estimated using the trained independent SVRs.

\section{Experimental Setup and Results}
\begin{figure}
  \includegraphics[width=\columnwidth]{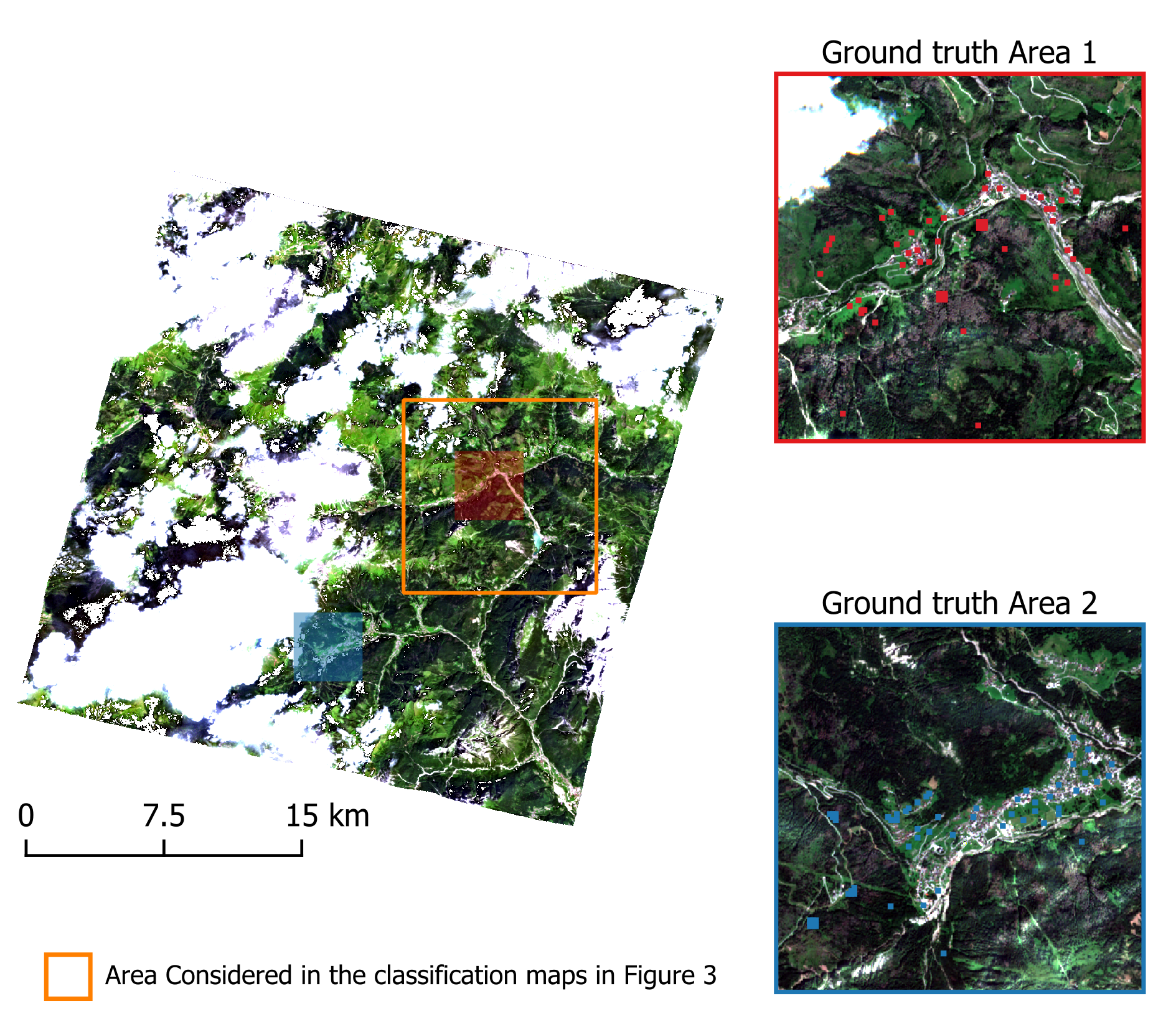}
  \caption{Illustration of the study area location and ground truth data.}
  \label{fig:scenario}
\end{figure}

To evaluate the effectiveness of the proposed methodology, we utilized a dataset made up of a hyperspectral image of the Dolomites region in northeastern Italy, (see Figure \ref{fig:scenario}). The image was acquired by the PRISMA satellite on August 7, 2024, with a spatial resolution of 30 meters. The image at processing level 2D  was co-registered using the state-of-the-art tool AROSICS \cite{scheffler2017arosics}. The dataset comprises ground truth data consisting of two types of samples. Pure healthy trees samples are obtained through photointerpretation. Mixed samples are collected through ground surveys conducted in late summer 2024 and include detailed information on the total number of trees, the number of trees attacked by bark beetle, and the number of dead trees within 30 × 30 meter areas. These details are processed to compute the abundance (percentage coverage) for each pixel corresponding to healthy, affected, and dead trees.

The labeled samples are distributed across two distinct areas in the hyperspectral image. The pure samples were randomly distributed, whereas the mixed samples were organized into three contiguous blocks in each area. For training and validation purposes, the samples within each area were split into 70\% for training and 30\% for validation, and to ensure reliable experimental results a 5-fold cross-validation strategy was utilized.

Given the peculiarity of this work that uses satellite hyperspectral images, particularly those from PRISMA, it is interesting to understand their general performance in detecting bark beetle infestations, comparing it with the results obtained using satellite multispectral Sentinel-2 data.
For this comparison, we selected an image acquired on August 15, 2024, as it is the closest date to the PRISMA image acquisition without significant cloud cover over the study area. The Sentinel-2 image is resampled to a 30-meter resolution using cubic interpolation. To match the ground truth data with the imagery, we identify the pixels from the PRISMA and Sentinel-2 images that encompassed the majority of each 30 × 30 meter ground truth area.

To assess the quality of the features extracted by the proposed methodology, we compare its performance against the direct use of all the original spectral bands of PRISMA and Sentinel-2 as input to the independent support vector regressor (SVR) estimators, one for each class, used as explained in the methodology section.

To ensure that the results are independent of hyperparameter tuning, we optimize the hyperparameters of the proposed model using the Tree-structured Parzen Estimator (TPE) algorithm \cite{bergstra2011algorithms}, while maintaining a fixed hyperparameter configuration for the independent SVR estimators, tuned using a grid search strategy for the regularization parameter C and the RBF kernel parameter $\sigma$. The optimal hyperparameter configuration identified is as follows: $\textit{lr}_{\text{self}}$ = 0.0094, $\textit{wd}_{\text{self}}$ = 0.0343, $\textit{lr}_{\text{ft}}$ = 0.0051, $\textit{wd}_{\text{ft}}$ = 0.0066 and $\tau$ = 0.0866.

The quantitative results are evaluated as the average Root Mean Squared Error (RMSE) across the five folds for the classes related to bark beetle attacks (“Healthy”, “Affected” and “Dead”), with the average RMSE for all classes presented in Table \ref{tab:res}.

The results clearly demonstrate that the features extracted by our model, when used as input to the SVR estimators, outperform the original spectral bands, achieving an average RMSE of 0.1029 across the three classes, which is significantly lower than those of the other two approaches (0.1770 for Sentinel-2 and 0.1353 for PRISMA with SVR). Furthermore, the lower RMSE observed for the PRISMA data suggests that the higher spectral resolution and the larger number of bands in PRISMA, compared to Sentinel-2, provide a significant advantage in the detection of bark beetle attacks.

\definecolor{Mercury}{rgb}{0.901,0.901,0.901}
\begin{table}
\centering
\caption{Average RMSE across the folds of the three different approaches.}
\label{tab:res}
\begin{tblr}{
  width = \linewidth,
  colspec = {Q[181]Q[230]Q[142]Q[156]Q[123]Q[152]},
  cell{4}{3} = {Mercury},
  cell{4}{4} = {Mercury},
  cell{4}{5} = {Mercury},
  cell{4}{6} = {Mercury},
  vline{3} = {1-4}{},
  vline{6} = {1-4}{},
  hline{1,5} = {-}{0.08em},
  hline{2} = {-}{0.05em},
}
\textbf{Data} & \textbf{Feature} & \textbf{Healthy} & \textbf{Affected} & \textbf{Dead} & \textbf{Average}\\
S2 & Original ch. & 0.2254 & 0.1320 & 0.1734 & 0.1770\\
PRISMA & Original ch. & 0.1538 & 0.1076 & 0.1447 & 0.1353\\
PRISMA & Model & \textbf{0.1091} & \textbf{0.0811} & \textbf{0.1186} & \textbf{0.1029}
\end{tblr}
\end{table}

\begin{figure}
  \includegraphics[width=\columnwidth]{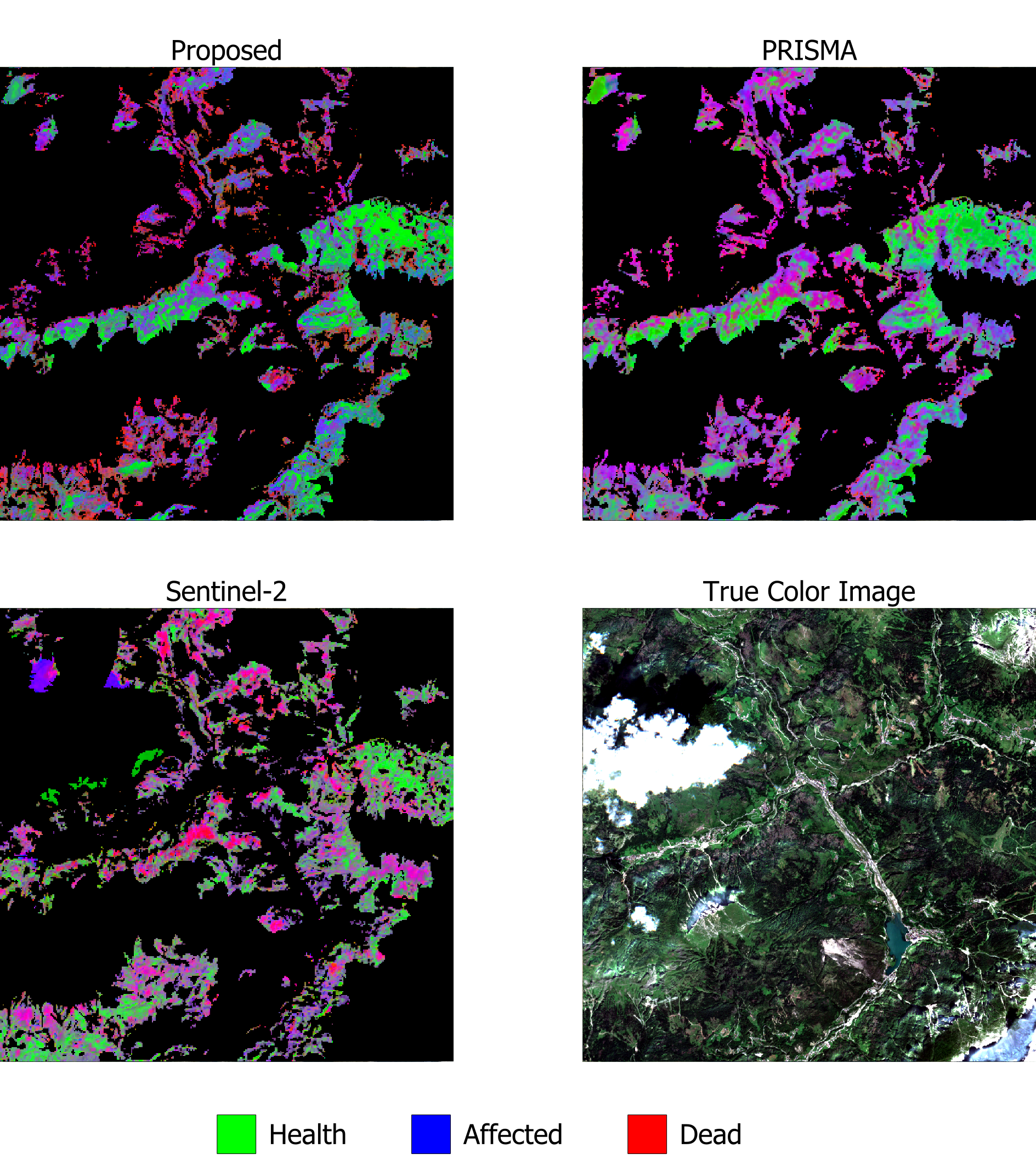}
  \caption{Portion of the classification maps visualized by normalizing each individual RGB channel, corresponding to the three considered classes, to the minimum and maximum values across the maps. The resulting colors reflect the varying abundances of each of the three classes for every pixel.}
    \label{fig:clmap}
\end{figure}

The classification maps, a part of which is presented in Figure \ref{fig:clmap}, are generated using the Coniferous map provided by the Veneto Region to mask areas where Norway Spruce are present and so where bark beetle attacks may occur. 

The classification map generated by the proposed feature extraction method, when compared to the one derived from all original PRISMA bands directly given as input to the SVR estimators, demonstrates a more accurate estimation of abundances, and an effective managing of the spectral variability in the image. In areas unaffected by bark beetle infestations, the proposed method results in the detection of a higher proportion of healthy trees. Similarly, in regions where dead trees predominate, our method produces more accurate results. When compared to methods using Sentinel-2 data, the PRISMA-based approach shows significantly higher accuracy in detecting bark beetle outbreaks. This improved accuracy is particularly noteworthy in areas with a high degree of class mixing, where the higher spectral resolution enables the identification of finer details in the data, ultimately leading to a more precise estimation of abundances.

In summary, the enhanced accuracy in abundance estimation achieved by our proposed method, coupled with the overall superior qualitative results, justifies the adoption of the proposed DL-based feature extraction approach.
\section{Conclusions}

This paper has presented a novel few-shot learning approach to the sub-pixel detection of bark beetle infestations in PRISMA hyperspectral data. By leveraging contrastive learning and SVRs, the proposed method effectively addresses the challenges of limited labeled data and the relatively low spatial resolution of hyperspectral satellite imagery. The use of a pre-trained CNN encoder allows for robust feature extraction from hyperspectral data, while fine-tuning enables adaptation to the specific task of bark beetle detection.

Our results demonstrate the efficacy of the proposed method in detecting bark beetle infestations, showing both quantitative and qualitative improvements over the use of solely spectral bands of PRISMA and Sentinel-2. The proposed method achieved a lower average RMSE across the three classes, significantly outperforming the other two approaches. Additionally, PRISMA data outperformed Sentinel-2 in detection accuracy, indicating the advantages of its higher spectral resolution and number of spectral bands.

In general, the combination of contrastive learning with few-shot learning techniques enables efficient extraction of information from minimal labeled data, while satellite hyperspectral imagery offers a scalable solution for large-scale forest health monitoring compared to UAV or airborne platforms.

Future work will focus on refining the accuracy and generalization capacity of the model to a broader range of ecological conditions, as well as exploring the integration of other hyperspectral remote sensing platforms. Furthermore, the model can be tested on different bark beetle infestation tasks or, more generally, on other forest health detection tasks to evaluate its robustness and adaptability.

\small
\bibliographystyle{IEEEtranN}
\bibliography{references}

\end{document}